\documentclass[11pt]{article}
\usepackage[utf8]{inputenc}
\usepackage{amsmath}
\usepackage{amssymb}
\usepackage{amsfonts}
\usepackage{graphicx}
\usepackage{geometry}
\usepackage{booktabs}
\usepackage{caption}
\usepackage{subcaption}
\usepackage{hyperref}
\usepackage{authblk}
\usepackage{float} 
\usepackage{tikz}
\usetikzlibrary{shapes.geometric, arrows.meta, positioning}
\usepackage{lmodern}
\PassOptionsToPackage{x11names, svgnames, table}{xcolor}
\usepackage[x11names,svgnames, table]{xcolor} 
\definecolor{myblue}{HTML}{799bc8}   

\hypersetup{
    colorlinks=true,
    linkcolor=blue,
    filecolor=magenta,
    urlcolor=cyan,
}

\title{uGMM-NN: Univariate Gaussian Mixture Model Neural Network}
\author{
    Zakeria Sharif Ali \\
    Independent Researcher \\
    \texttt{z@zakeria.com} 
}

\date{}

\begin{document}

\maketitle

\begin{abstract}
This paper introduces the Univariate Gaussian Mixture Model Neural Network (uGMM-NN), a novel architecture that embeds probabilistic reasoning directly into deep networks. Unlike traditional neurons, which apply weighted sums followed by fixed nonlinearities, each node within the uGMM-NN is parameterized by a univariate Gaussian mixture, with learnable means, variances, and mixing coefficients.
This design enables richer representations by capturing multimodality and uncertainty at the level of individual neurons, while retaining the scalability of standard feedforward networks. 
We demonstrate that uGMM-NN can achieve competitive discriminative performance compared to conventional multilayer perceptrons, while additionally offering a probabilistic interpretation of activations. 
The proposed framework provides a foundation for integrating uncertainty-aware components into modern neural architectures, opening new directions for both discriminative and generative modeling.
\end{abstract}
\noindent \textbf{Keywords:} probabilistic neurons, deep learning, probabilistic circuits, Gaussian mixture, uGMM Layer,  uGMM, uGMM-NN

\section{Introduction}
Deep neural networks have transformed machine learning, excelling in tasks such as image classification and natural language processing through hierarchical feature learning \cite{Krizhevsky2012}. However, traditional neurons, which compute deterministic weighted sums followed by nonlinear activations (e.g., ReLU, sigmoid), struggle to model uncertainty or multimodal distributions prevalent in real-world data.
This limitation has historically been addressed by probabilistic graphical models, such as Bayesian Networks \cite{Pearl1988} and Markov Random Fields \cite{Kindermann1980}, which offer robust frameworks for uncertainty quantification and complex dependency modeling \cite{Koller2009}. These models provide a strong conceptual foundation, but often lack the deep hierarchical feature learning capabilities of modern neural networks. \\ 

A key research focus has therefore been on bridging the gap between these two paradigms. This has led to approaches that incorporate the probabilistic principles of graphical models directly into deep learning architectures. For example, Bayesian Neural Networks (BNNs) embed uncertainty into network weights \cite{Gal2016}, while Probabilistic Circuits (PCs), including Sum-Product Networks (SPNs) \cite{Poon2011SPN, Choi2020PC, Peharz2020Einsum}, are deep probabilistic models that build on a formal probabilistic structure, fusing the representational power of graphical models with the hierarchical feature learning of neural networks. \\

In contrast, this paper introduces a novel approach by embedding a univariate Gaussian Mixture Model (uGMM) directly into the network’s computational units, enabling each neuron to represent a multimodal univariate Gaussian distribution, thereby directly addressing the limitations of deterministic neurons from a new perspective. While our primary focus is on discriminative applications
where the goal is to predict a class label, the probabilistic backbone of our model shares conceptual similarities with probabilistic circuits. We evaluate the uGMM-NN across two representative datasets and two architectural paradigms:
\begin{enumerate}
\item On MNIST, we perform a direct comparison of discriminative performance between the uGMM-NN and a standard Multi-Layer Perceptron (MLP), and further demonstrate the versatility of our approach by replacing the fully connected layers of a convolutional network with uGMM neurons (CNN-uGMM vs. CNN-MLP).
\item On the classic Iris dataset, we offer a preliminary demonstration of the model's capacity for generative modeling of a joint distribution. This experiment serves to illustrate the conceptual connection between uGMM-NN and probabilistic graphical models.
\end{enumerate}

\section{Related Work}
uGMM-NN relates to several paradigms in probabilistic graphical modelling and deep probabilistic (differentiable) learning. Bayesian Neural Networks (BNNs) model uncertainty over weights \cite{BNN}, providing a principled route to calibrated predictions, though often at the cost of heavier inference. Probabilistic Circuits, such as Sum-Product Networks \cite{Poon2011SPN}, enable tractable inference for complex distributions \cite{Choi2020PC, Peharz2020Einsum}. These models provide a framework for explicitly representing probability distributions with guarantees on tractability, a concept uGMM-NN draws inspiration from by embedding a tractable probabilistic model within each neuron. \\

In parallel, Kolmogorov–Arnold Networks (KAN) \cite{KAN} pursue interpretability and flexibility by replacing the linear affine maps of MLPs with learnable univariate functions along edges. KAN thus substitutes the standard transformation: \[
    x \mapsto W x + b
\] 
with a composition of spline-like functions, yielding more transparent functional dependencies 
than opaque weight matrices. uGMM-NN makes an analogous replacement at the unit level: 
instead of a linear neuron followed by a pointwise nonlinearity, each computational unit is a univariate Gaussian Mixture Model (uGMM) that outputs responsibilities and mixture statistics.
uGMM-NN swaps linear neurons for probabilistic neurons, yielding interpretable, multimodal, and uncertainty-aware intermediate representations. \\

This substitution brings two benefits central to our aims. First, interpretability: mixture components act as localized, semantically meaningful “submodes,” exposing how a unit partitions its input space, akin to KAN’s readable univariate maps, but now in explicitly probabilistic terms. Second, expressivity under uncertainty: mixtures capture multimodality and uncertainty within each unit, rather than deferring uncertainty to the weight space (as in BNNs) or to global circuit structure (as in SPNs and probabilistic circuits).  In short, like KAN replaces linear MLP layers with functional mappings, uGMM-NN replaces them with probabilistic mixture units, yielding a network whose intermediate computations are distributions, interpretable, multimodal, and uncertainty-aware rather than single deterministic activations.\\

Radial Basis Function (RBF) networks also employ Gaussian-shaped responses at the unit level \cite{rbf_orig}, but they are not probabilistic models: their activations are unnormalized basis functions centered at fixed prototypes, without mixture weights that sum to one or a likelihood-based interpretation. By contrast, each uGMM neuron defines a normalized Gaussian mixture with trainable means, variances, and mixing weights, and outputs a log-density. Consequently, despite a superficial similarity in functional form, uGMM-NN is conceptually closer to probabilistic circuits than to classical RBF networks.

\section{uGMM-NN}
The uGMM-NN replaces standard MLP neurons with probabilistic units, each modeling a univariate Gaussian mixture over its inputs. This design enables the network to capture multimodal behavior and input uncertainty at the level of individual neurons, while remaining fully differentiable and trainable with standard supervised losses. Conceptually, each uGMM neuron serves as a direct replacement for a traditional linear neuron followed by a pointwise nonlinearity, providing richer intermediate representations.

\subsection{Univariate GMM Nodes}\label{sec:ugmm}

In traditional multilayer perceptrons (MLPs), each neuron computes a weighted sum of inputs followed by a nonlinear activation function such as ReLU or sigmoid. These scalar activations are then passed forward to the next layer. In contrast, each neuron in uGMM-NN computes the log-density of a univariate Gaussian Mixture Model (uGMM), making the neuron's output a probabilistic quantity rather than a deterministic scalar.

Let $\mathbf{x} = [x_1, \dots, x_N]$ denote the inputs to neuron $j$ from the previous layer, 
where $N$ is the number of neurons in that layer. In uGMM-NN, each neuron $j$ defines a univariate Gaussian mixture over a latent variable $y$, parameterized by learnable means $\mu_{j,k}$, variances 
$\sigma^2_{j,k}$, and mixing weights $w_{j,k}$ for each input $x_k$:

\[
P_j(y) = \sum_{k=1}^{N} w_{j,k} \, \mathcal{N}(y \mid \mu_{j,k}, \sigma^2_{j,k}),
\quad \text{with} \quad \sum_{k=1}^{N} w_{j,k} = 1.
\]

Each input $x_k$ contributes one component to the mixture. Therefore, the number of mixture 
components $N$ is determined by the width of the previous layer, rather than chosen as a separate 
hyperparameter.\\

Expanding the Gaussian explicitly, the probability density becomes:

\[
P_j(y) = \sum_{k=1}^{N} w_{j,k} \cdot \frac{1}{\sqrt{2\pi \sigma^2_{j,k}}} \exp\left( -\frac{(y - \mu_{j,k})^2}{2\sigma^2_{j,k}} \right).
\]

The neuron's output is then defined as the log-density:

\[
\log P_j(y) = \log \left( \sum_{k=1}^{N} \exp\left[
    \log w_{j,k}
    - \frac{1}{2} \log(2\pi \sigma^2_{j,k})
    - \frac{(y - \mu_{j,k})^2}{2\sigma^2_{j,k}}
\right] \right),
\]

which is computed using the log-sum-exp trick for numerical stability. \\

We refer to this log-density output $\log P_j(y)$ as the neuron's \emph{activation}\footnote{In standard neural networks, an “activation” refers to a scalar value resulting from a nonlinear transformation of a weighted sum. In uGMM-NN, each neuron instead outputs the log-density of a mixture distribution, which we treat as the probabilistic analogue of a traditional activation. These log-densities are passed to the next layer, participate in gradient flow, and play the same architectural role.}. This probabilistic activation is propagated forward through the network and forms the basis for both classification and likelihood-based training. 
At subsequent layers, the inputs to each uGMM neuron are the log-density activations produced by the preceding layer, ensuring that a consistent probabilistic representation is maintained throughout the network.

\begin{figure}[H]
\centering
\begin{tikzpicture}[
    neuron/.style={circle, draw, fill=myblue, text=white, minimum size=20mm, inner sep=0pt},
    input/.style={circle, draw, fill=black, text=white, minimum size=10mm, inner sep=0pt},
    arrow/.style={-Stealth, thick},
    label/.style={font=\small}
]
\node[input] (x1) at (0,2) {\(x_1\)};
\node[input] (x2) at (0,0) {\(x_2\)};
\node[draw=none] (xdots) at (0,-1) {\(\vdots\)};
\node[input] (xn) at (0,-2) {\(x_N\)};

\node[neuron] (n) at (6,0) {Neuron \(j\)};

\node[draw=none] (output) at (9,0) {\(\log P_j(y)\)};

\draw[arrow] (x1) -- (n) node[midway, above, yshift=2mm, rotate=-5] {\(\mu_{j,1}, \sigma_{j,1}, w_{j,1}\)};
\draw[arrow] (x2) -- (n) node[midway, above, yshift=0mm] {\(\mu_{j,2}, \sigma_{j,2}, w_{j,2}\)};
\draw[arrow] (xn) -- (n) node[midway, above, yshift=0mm, rotate=12] {\(\mu_{j,N}, \sigma_{j,N}, w_{j,N}\)};
\draw[arrow] (n) -- (output);
\end{tikzpicture}

\caption{Schematic of a uGMM-NN neuron. Each neuron models a uGMM with \(N\) components based on inputs from the previous layer, with each connection parameterized by mean (\(\mu_{j,k}\)), standard deviation (\(\sigma_{j,k}\)), and mixing weight (\(w_{j,k}\)), outputting the log probability \(\log P_j(y)\).}
\label{fig:ugmmnn}
\end{figure}

\subsection{Probabilistic Interpretability of Neurons}

Unlike standard MLP neurons, which produce deterministic scalar activations, uGMM-NN neurons output the log-density of a univariate Gaussian mixture. This probabilistic quantity not only propagates forward through the network but also encodes interpretable structure at the unit level.\\

Figure~\ref{fig:ugmm_mixture} illustrates a single uGMM neuron with three mixture components. Each component is parameterized by a mean $\mu_{j,k}$, standard deviation $\sigma_{j,k}$, and mixing weight $w_{j,k}$ (with the variance $\sigma_{j,k}^2$ used in the log-density computation). Together, these define the neuron's overall probability density \(P_j(y)\), with the dashed black line representing the full mixture. The neuron’s output is the log-density $\log P_j(y)$, which we interpret as its activation.\\

This formulation enables an intuitive understanding of what the neuron "detects." For example, a component with a high mixing weight $w_{j,k}$ indicates that the neuron assigns significant probability mass to the corresponding region of its input space. Similarly, the spread of the Gaussian (via $\sigma_{j,k}^2$) reflects how confidently the neuron attributes input mass to a given region of its input space.
From this perspective, each neuron defines a probabilistic partition over its input domain, where mixture components correspond to interpretable submodes of the neuron's response. This is in contrast to traditional activations, which typically lack such localized semantics. As a result, uGMM-NN offers built-in interpretability, where the internal representation of each layer is not just a scalar value but a structured probability landscape.\\

Figure~\ref{fig:ugmmnn} (introduced in the previous section) situates this mechanism within the full network, illustrating how log-densities are propagated forward through successive layers. Together, Figures~\ref{fig:ugmmnn} and~\ref{fig:ugmm_mixture} demonstrate how probabilistic activations are formed and refined across the network. This property is particularly advantageous in applications such as healthcare or autonomous systems, where interpretability and uncertainty quantification are essential.

\begin{figure}[H]
\centering
\begin{tikzpicture}[scale=1.0]
\draw[->] (-0.5,0) -- (5.5,0) node[right] {$y$};
\draw[->] (0,-0.5) -- (0,3) node[above] {Density};

\draw[blue, thick, domain=0:5, samples=200] plot (\x,{2*exp(-0.5*((\x-1.5)/0.5)^2)});
\node[blue] at (1.5,2.2) {$\mu_{j,1}$};

\draw[red, thick, domain=0:5, samples=200] plot (\x,{1.5*exp(-0.5*((\x-3)/0.3)^2)});
\node[red] at (3,1.7) {$\mu_{j,2}$};

\draw[green!70!black, thick, domain=0:5, samples=200] plot (\x,{1*exp(-0.5*((\x-4)/0.4)^2)});
\node[green!70!black] at (4,1.1) {$\mu_{j,3}$};

\draw[black, dashed, thick, domain=0:5, samples=200] plot (\x,{2*exp(-0.5*((\x-1.5)/0.5)^2)+1.5*exp(-0.5*((\x-3)/0.3)^2)+1*exp(-0.5*((\x-4)/0.4)^2)});
\node[black] at (5,2.8) {$P_j(y)$};
\end{tikzpicture}
\caption{Illustration of a single uGMM neuron with three univariate Gaussian mixture components. The means (\(\mu_{j,k}\)), variances (\(\sigma_{j,k}^2\)), and mixing weights (\(w_{j,k}\)) define the neuron's probability density \(P_j(y)\). The dashed black line shows the combined mixture density.}
\label{fig:ugmm_mixture}
\end{figure}

\subsection{Network Structure and Training}
The uGMM-NN is organized in a standard feedforward manner with input, hidden, and output layers. Each hidden layer can be interpreted as a Gaussian mixture layer, where every neuron computes a univariate Gaussian mixture model over the activations of the previous layer, producing log-densities that serve as inputs for the next layer. Importantly, the number of mixture components in each neuron is automatically determined by the width of the preceding layer, ensuring a direct mapping from inputs to probabilistic components.\\

Stacking multiple univariate Gaussian Mixture Model (uGMM) layers allows the network to build hierarchical probabilistic representations, capturing increasingly complex and multimodal patterns in the data. Unlike traditional MLPs, where each neuron performs a 1D affine transformation followed by a nonlinearity and outputs a deterministic scalar, uGMM neurons operate on univariate inputs and output log-probability values under a learned Gaussian mixture. These can be viewed as 1D probabilistic operations, analogous in shape and dimension to their MLP counterparts. By focusing on univariate GMMs at each neuron, uGMM-NN avoids the exponential complexity often associated with multivariate mixture models, allowing it to scale efficiently to deep architectures. This architectural simplicity enables practical training and inference on large datasets while retaining the expressive power of probabilistic modeling.

For discriminative tasks, the output (root) layer log-densities can be converted into class probabilities using the softmax function, after which training minimizes the standard cross-entropy loss. \\

Gradients are backpropagated through the log-density computations using automatic differentiation, and training is performed with gradient-based optimizers such as Adam or SGD. To improve numerical stability, the log-sum-exp trick is employed when computing gradients through the mixture sums, preventing overflow or underflow in deep networks. \\

Despite the added complexity of per-neuron mixture computations, uGMM layers are naturally amenable to GPU parallelization. Each neuron's responsibilities and log-density contributions are computed using vectorized tensor operations over batches, neurons, and mixture components. Specifically, the core computation, which evaluates the log-density of each input under a learnable mixture of Gaussians, is implemented as a fully batched operation using PyTorch, with dropout applied directly to the mixture components during training. The use of \texttt{logsumexp} ensures numerical stability, while broadcasting enables efficient computation across batch dimensions. As a result, uGMM-NN achieves training and inference speeds comparable to standard multilayer perceptrons despite the probabilistic overhead \cite{ugmmnn_zenodo, ugmmnn_github}.

\subsection{Dropout in uGMM-NN}
To improve generalization and prevent overfitting, dropout can be applied to uGMM neurons in a manner analogous to standard MLP layers. 
Specifically, a Bernoulli mask is applied independently to each mixture component \(k\) of neuron \(j\):
\[
d_{j,k} \sim \text{Bernoulli}(1-p),
\]
where \(p\) is the dropout probability. Components that are "dropped" are effectively removed from the mixture by setting their log-density contribution to \(-\infty\). The resulting log-density with dropout is:
\[
\log P_j(y)_{\text{dropout}} = \log \sum_{k=1}^{N} w_{j,k} \, d_{j,k} \, \mathcal{N}(y \mid \mu_{j,k}, \sigma_{j,k}^2),
\]
where \(w_{j,k}\), \(\mu_{j,k}\), and \(\sigma_{j,k}^2\) are the mixture weights, means, and variances of the \(k\)-th component of neuron \(j\), respectively. Dropout is applied only during training and is disabled during inference, allowing the full mixture to contribute to the neuron's output.

\subsection{Universal Approximation Perspective}
Each uGMM neuron outputs a log-density defined by a mixture of univariate Gaussians. Since a mixture of Gaussians can approximate any smooth univariate probability density on a compact domain to arbitrary accuracy, uGMM neurons are, in principle, universal approximators of univariate densities. \\

When stacked in a feedforward network, the resulting uGMM-NN can approximate complex multivariate distributions over the inputs. For discriminative tasks, this implies that the network can represent any conditional probability distribution $P(y \mid x)$ given sufficient width (number of neurons per layer) and depth (number of layers). \\

In comparison, standard MLPs and Kolmogorov–Arnold Networks (KANs) provide universal approximation guarantees for deterministic functions. uGMM-NN extends this notion to probabilistic, multimodal representations, combining expressive power with interpretable uncertainty through mixture parameters $(w_{j,k}, \mu_{j,k}, \sigma_{j,k}^2)$.

\medskip
\noindent\hrulefill
\medskip

The uGMM-NN architecture and all experiments were implemented in PyTorch. 
For full reproducibility, the code corresponding to this work is available via a Zenodo release (initially published on July 20, 2025) \cite{ugmmnn_zenodo}, 
and the latest development version can be accessed on GitHub \cite{ugmmnn_github}.

\section{Experiments}
We conduct a series of experiments to evaluate the performance and properties of uGMM-NN, focusing on both generative and discriminative tasks, and comparing it to standard MLP architectures. Specifically, we perform the following experiments:

\begin{enumerate}
    \item \textbf{Generative Learning on Iris:}
    A proof-of-concept experiment where the uGMM-NN is trained to model the joint probability distribution of the features and labels, demonstrating its ability to capture multimodal distributions and perform simple posterior inference.

    \item \textbf{Discriminative Comparison on MNIST:}
    A direct comparison between a standard fully-connected MLP and a uGMM-NN of equivalent depth and hidden size, evaluating classification accuracy and training stability on a classic image classification benchmark.

    \item \textbf{Convolutional Architectures:}
    Extending the comparison to convolutional networks, we replace the dense layers in a CNN with uGMM neurons (CNN-uGMM) and compare performance against a conventional CNN with standard MLP layers (CNN-MLP). This demonstrates the versatility of uGMM neurons in modern architectures.
\end{enumerate}
In the MNIST classification experiments, dropout is applied consistently to prevent overfitting, and hyperparameters are selected based on validation performance to ensure fair comparison between models.

\subsection{Datasets}
We considered two datasets for our evaluation: 
\begin{itemize}
    \item \textbf{MNIST} \cite{lecun1998mnist}: A common benchmark for deep learning image classification, consisting of 70,000 grayscale images (60,000 train, 10,000 test) of handwritten digits (28$\times$28 pixels), with 10 output classes. 
    \item \textbf{Iris} \cite{fisher1936iris}: A classic small-scale dataset often employed for foundational classification tasks, comprising 150 samples with 4 numerical features and 3 output classes, corresponding to different iris species. 
\end{itemize}

\subsection{Training Setup}
This section details the training methodologies and hyperparameters used for our models.

\subsubsection{Generative Training}
For the generative experiment on the Iris dataset, the network was trained to model the joint probability distribution of the features and labels, $P(\mathbf{x}, y)$. The architecture consisted of two simple hidden layers with 10 and 5 uGMM neurons, respectively. The network outputs a single log-density value, representing the log-likelihood of the input.

The model was trained to maximise the average log-likelihood of the training data, using the Adam optimizer with a learning rate of $10^{-3}$. No dropout was applied, and training proceeded for 300 epochs to ensure convergence.

\subsubsection{Discriminative Training}

In the MNIST classification experiments, all models were trained using the Adam optimizer with default parameters ($\beta_1 = 0.9, \beta_2 = 0.999$) and a cross-entropy loss function. Training was performed for 100 epochs using the $\texttt{MultiStepLR}$ learning rate scheduler with a decay factor of $\gamma = 0.1$. \\

For the fully connected setting, we compared a standard MLP with a uGMM-NN of equivalent depth and hidden layer sizes. For the uGMM-NN, mixture parameters $(\mu_{j,k}, \sigma_{j,k}^2, w_{j,k})$ were initialized randomly, and dropout was applied consistently to prevent overfitting. Hyperparameters, including dropout probability and learning rate schedule milestones, were selected based on validation performance to ensure a fair comparison between models. \\

Similarly, for the convolutional models, we trained a conventional CNN with MLP layers (CNN-MLP) and a CNN where the dense layers were replaced with uGMM neurons (CNN-uGMM). Dropout and other hyperparameters were chosen analogously to the fully-connected setting to maintain a consistent training protocol across architectures. Table~\ref{tab:training_params} summarizes the optimized hyperparameters for all MNIST discriminative models.

\begin{table}[H]
\centering
\caption{Optimized Training hyperparameters for MNIST Discriminative Models}
\vspace{0.5em}
\begin{tabular}{lccc}
\toprule
\textbf{Model} & \textbf{Initial Learning Rate} & \textbf{Dropout $p$ (Layer)} & \textbf{LR Milestones} \\
\midrule
MLP & $10^{-3}$ & $0.5$ (128-unit) & $[20, 45]$ \\
\textbf{uGMM-NN} & $10^{-2}$ & $0.5$ (128-unit) & $\mathbf{[40]}$ \\
CNN-MLP & $10^{-3}$ & $0.5$ (128-unit) & $[20, 45]$ \\
\textbf{CNN-uGMM} & $10^{-3}$ & $0.5$ (128-unit) & $\mathbf{[45, 80]}$ \\
\bottomrule
\end{tabular}
\label{tab:training_params}
\end{table}

The uGMM-NN utilised a larger initial learning rate and adjusted learning-rate milestones, reflecting the faster adaptation required by its mixture parameters. Moreover, a consistent dropout rate of $p = 0.5$ was applied across all models to ensure comparable regularisation.

\subsection{Performance}
When trained generatively, the uGMM-NN model achieved perfect classification accuracy on the Iris dataset. Predictions were obtained via posterior inference:
\[
\hat{y} = \arg\max_{c} P(y = c \mid \mathbf{x}) = \arg\max_{c} P(y = c, \mathbf{x}),
\]
where \(\mathbf{x}\) is the feature vector. This strong performance on a simple, foundational dataset demonstrates its capability to perform discriminative inference from a generatively trained probabilistic model. These results motivate future evaluation of uGMM-NN on more complex tabular benchmarks, though the primary focus of this paper remains on its discriminative abilities, particularly in comparison to traditional MLPs. \\

For the fully connected models on MNIST, both the MLP and the uGMM-NN converged to strong performance after 100 epochs. The MLP achieved a stable test accuracy of $98.15\%$ over the final training window (epochs~90--100), as reported in Table~\ref{tab:results}. The uGMM-NN initially reached a slightly lower but still competitive accuracy of $97.8\%$. This small gap is consistent with the wider dynamic range and higher variance of the log-density activations produced by uGMM neurons, which propagate richer statistical information than the deterministic scalar activations of the MLP. To explicitly address this effect and stabilise activation scales, \emph{Layer Normalization} ~\cite{ba2016layer} was applied to the uGMM-NN. Although Layer Normalization slowed the initial learning trajectory, it ultimately enabled the model to reach a higher and more stable accuracy of $98.34\%$. These results demonstrate that, when the statistical characteristics of its probabilistic activations are properly managed, the uGMM-NN can match or marginally exceed the MLP’s discriminative performance while additionally offering uncertainty quantification and improved interpretability. \\

For the convolutional models on \text{MNIST}, all architectures achieved strong performance, with test accuracies approaching or exceeding $99\%$. The baseline CNN-MLP reached a test accuracy of $99.35\%$, reflecting the well-established effectiveness of convolutional architectures on this dataset. \\

Replacing the final fully connected layers with uGMM neurons resulted in a CNN-uGMM model that achieved an accuracy of $98.77\%$. When the same normalization strategy described above was applied, the CNN-uGMM reached a higher accuracy of $99.12\%$, demonstrating that uGMM layers can be integrated into convolutional architectures while preserving competitive discriminative performance. Importantly, this substitution introduces a transparent, probabilistic, and structurally interpretable alternative to the standard dense layer. \\

\begin{figure*}[h!]
    \centering
    \begin{subfigure}[t]{0.48\textwidth}
        \centering
        \includegraphics[width=\linewidth]{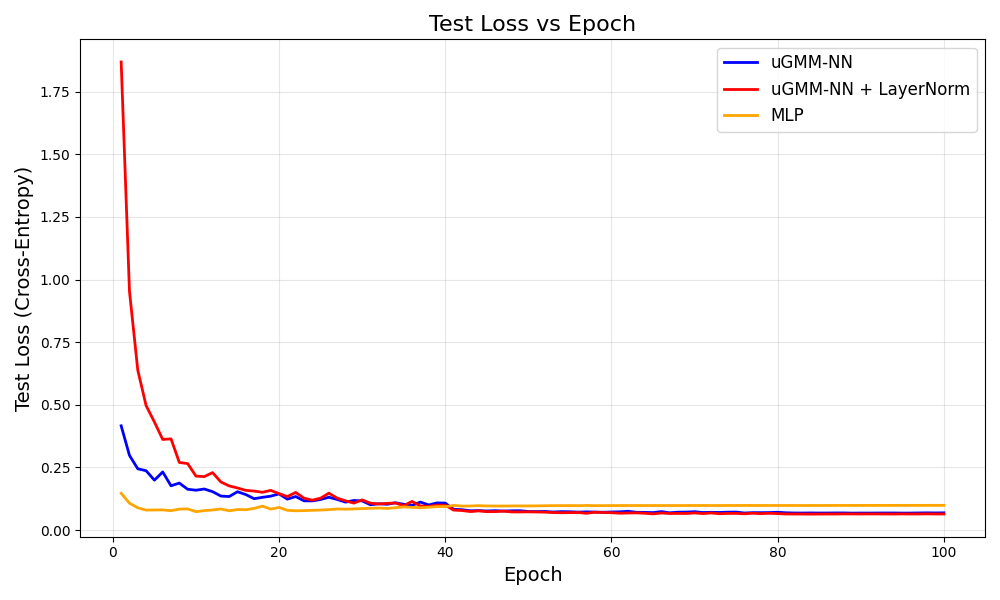}
        \caption{Fully-connected (uGMM-NN vs MLP) training on MNIST.}
        \label{fig:test_loss_ffnn}
    \end{subfigure}
    \hfill
    \begin{subfigure}[t]{0.48\textwidth}
        \centering
        \includegraphics[width=\linewidth]{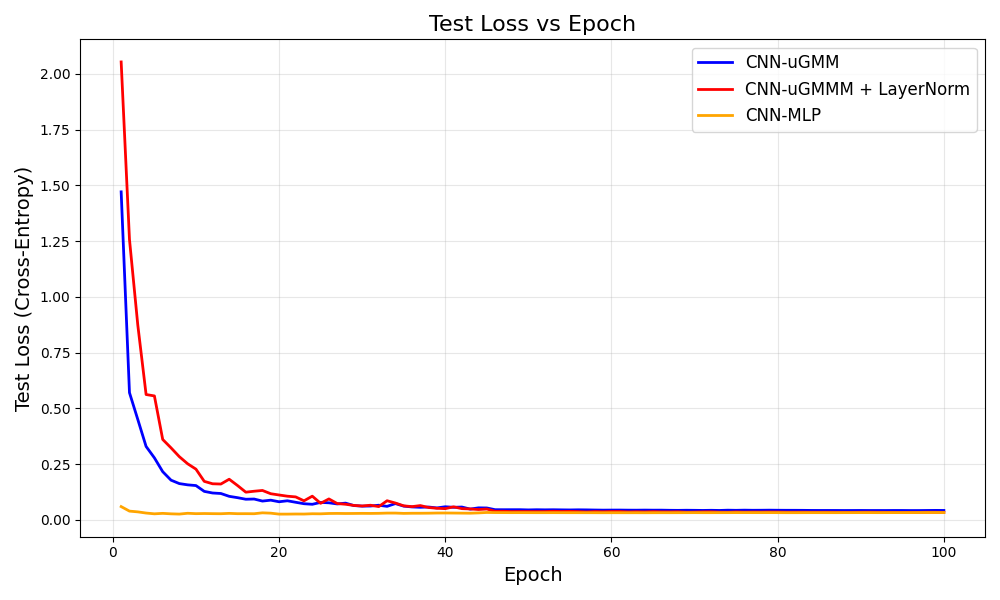}
        \caption{Convolutional (CNN-uGMM vs CNN-MLP) training on MNIST.}
        \label{fig:test_loss_cnn}
    \end{subfigure}
    \caption{
    Comparison of test loss convergence between MLP, uGMM-NN, and their LayerNorm-enhanced or CNN-augmented counterparts.
    The uGMM layers achieve comparable or superior final loss and stability, particularly when Layer Normalization is applied to stabilize the log-density activations.
    }
    \label{fig:test_loss_comparison}
\end{figure*}

Figure~\ref{fig:test_loss_comparison}(b) compares the test loss convergence of the CNN-uGMM and CNN-MLP architectures over 100 training epochs. Although final accuracy is comparable, the optimisation dynamics differ markedly. The deterministic CNN-MLP shows the expected smooth and rapid decline in cross-entropy loss, reaching a stable low value within the first 10 epochs. In contrast, the CNN-uGMM variants exhibit a steeper and more variable early trajectory, with substantially higher initial loss and pronounced oscillations during the first 20 epochs. This behaviour reflects the added stochasticity introduced by the uGMM log-density activations, which propagate distributional information rather than fixed point activations.

\begin{table}[H]
\centering
\caption{Test accuracy and training loss for uGMM-NN vs. MLP across benchmark datasets.}
\label{tab:results}
\begin{tabular}{llcc}
\toprule
\textbf{Dataset}          & \textbf{Model}    & \textbf{Test Accuracy (\%)} \\
\midrule
MNIST                    & MLP             & 98.15                          \\
MNIST                    & \textbf{uGMM-NN} & \textbf{97.8}                  \\
MNIST                    & \textbf{uGMM-NN + LayerNorm} & \textbf{98.34}      \\
MNIST                    & \textbf{CNN-uGMM} & \textbf{98.77}                \\
MNIST                    & \textbf{CNN-uGMM + LayerNorm} & \textbf{99.12}                \\
MNIST                    & CNN-MLP & 99.35                           \\
\bottomrule
\end{tabular}
\end{table}

\section{Discussion and Limitations}
The uGMM-NN integrates probabilistic modeling into deep learning, offering interpretable uncertainty quantification through its mixture components. Compared to standard MLP neurons, each uGMM neuron introduces additional parameters, specifically, means, variances, and mixing coefficients for each input, enabling richer, multimodal representations, albeit at the cost of increased model size. \\

Tying the means \((\mu_{j,k})\) of each Gaussian component to the corresponding input activation \((x_k)\), i.e., setting \(\mu_{j,k} = x_k\), is a straightforward way to reduce the number of learnable parameters. In this case, the neuron only needs to learn the mixing coefficients \((w_{j,k})\) and variances \((\sigma_{j,k}^2)\), simplifying the model while reducing expressivity, since the neuron can no longer shift submodes independently of the inputs. \\

A remaining limitation arises when using the model for generative inference, in a manner similar to Bayesian networks, or probabilistic circuits \cite{Choi2020PC,Peharz2020Einsum}. Computing the Most Probable Explanation (MPE) for a full generative network is non-trivial, and currently no efficient algorithm exists for uGMM-NN. Developing a tractable Viterbi-style procedure, typically based on a forward-backward pass, as used in Sum-Product Networks and Probabilistic Circuits \cite{Poon2011SPN, Peharz2020Einsum, Viterbi1967}, remains an open challenge for extending uGMM-NN to large-scale generative inference. However, unlike traditional graphical models, uGMM-NN avoids the need for complex Bayesian structure learning, which is generally intractable due to the combinatorial nature of graph search and the difficulty of marginal likelihood estimation. Instead, the architecture of a uGMM-NN is specified simply by choosing the number of layers and neurons, and can be trained end to end via standard backpropagation. This simplifies the model's design and eliminates the need for expert-guided structure discovery. Hence, while generative applications remain an interesting direction for future work, the primary contribution of uGMM-NN lies in its ability to replace deterministic neurons with probabilistic units for interpretable, discriminative inference.

\section{Conclusion and Future Work}
This paper introduced the Univariate Gaussian Mixture Model Neural Network (uGMM-NN), a novel feedforward architecture that integrates probabilistic reasoning directly into the computational units of neural networks. By embedding a univariate Gaussian mixture model (uGMM) into each neuron, uGMM-NN provides a richer parameterization of activations, enabling the network to capture multimodality and quantify uncertainty in ways that standard feedforward layers cannot. This combination of predictive capability and probabilistic interpretability makes uGMM-NN a promising building block for complex neural architectures. \\

Our preliminary experiments demonstrate that uGMM-NN achieves competitive performance compared to standard feedforward neural networks (FFNNs) when trained discriminatively. These results highlight the feasibility of replacing conventional neurons with mixture-based alternatives, opening a new design space for deep learning architectures. Importantly, the probabilistic structure of uGMM neurons also exposes introspective signals, such as uncertainty and mixture spread, which could enable higher-level reflection and adaptive learning mechanisms. \\

Future work will focus on several directions:
\begin{itemize}
\item Extending uGMM neurons to other neural architectures such as Recurrent Neural Networks (RNNs) \cite{hochreiter1997long} and Transformers \cite{Vaswani2017}, to test their versatility across sequential and attention-based domains.

\item Developing efficient Most Probable Explanation (MPE) inference algorithms, including Viterbi-style procedures, to make uGMM-NN scalable for generative applications \cite{Viterbi1967}.
\item Scaling experiments to larger and multimodal datasets to evaluate robustness, generalizability, and performance under more complex real-world conditions.
\item Investigating \emph{sparse uGMM layers}, in which each neuron maintains a controlled subset of connections rather than using the full \(N\)-component expansion of the current design, or expanding across additional dimensions to add additional mixture components. In particular, combining sparse uGMM layers with architectures such as Transformers could enable attention-guided probabilistic representations, selectively propagating distributional information along the most relevant connections while reducing computational overhead.
\end{itemize}

In summary, uGMM-NN provides a compelling direction for embedding probabilistic modeling into deep learning architectures. With further development of inference algorithms, scalability, and theoretical analysis, uGMM-NN has the potential to bridge the gap between discriminative neural networks and generative probabilistic models, enabling architectures that are both accurate and interpretable.

\section*{References}
\begingroup
\renewcommand{\section}[2]{}

\endgroup
\end{document}